\documentclass[10pt,twocolumn,letterpaper]{article}

\usepackage{iccv}              
\usepackage{caption}
\usepackage{amsmath}
\usepackage[ruled,vlined]{algorithm2e}
\usepackage{multirow}
\usepackage{colortbl}
\usepackage{xcolor}
\usepackage{booktabs}

\usepackage{cuted}

\definecolor{iccvblue}{rgb}{0.21,0.49,0.74}
\usepackage[pagebackref,breaklinks,colorlinks,allcolors=iccvblue]{hyperref}

\def\paperID{****} 
\def\confName{ICCV}
\def\confYear{2025}

\title{Long-Video Audio Synthesis with Multi-Agent Collaboration}

\author{
Yehang Zhang$^{1*}$ \quad Xinli Xu$^{1*}$ \quad Xiaojie Xu$^{1*}$ \quad Li Liu $^{1,2\dagger}$ \quad Ying-Cong Chen$^{1,2\dagger}$\\
$^1$HKUST(GZ)  \quad $^2$HKUST
\thanks{Equal contribution. $^{\dagger}$ Co-corresponding author.}
}

\begin{document}
\maketitle
\begin{strip}
    \centering
\includegraphics[width=0.8\textwidth]{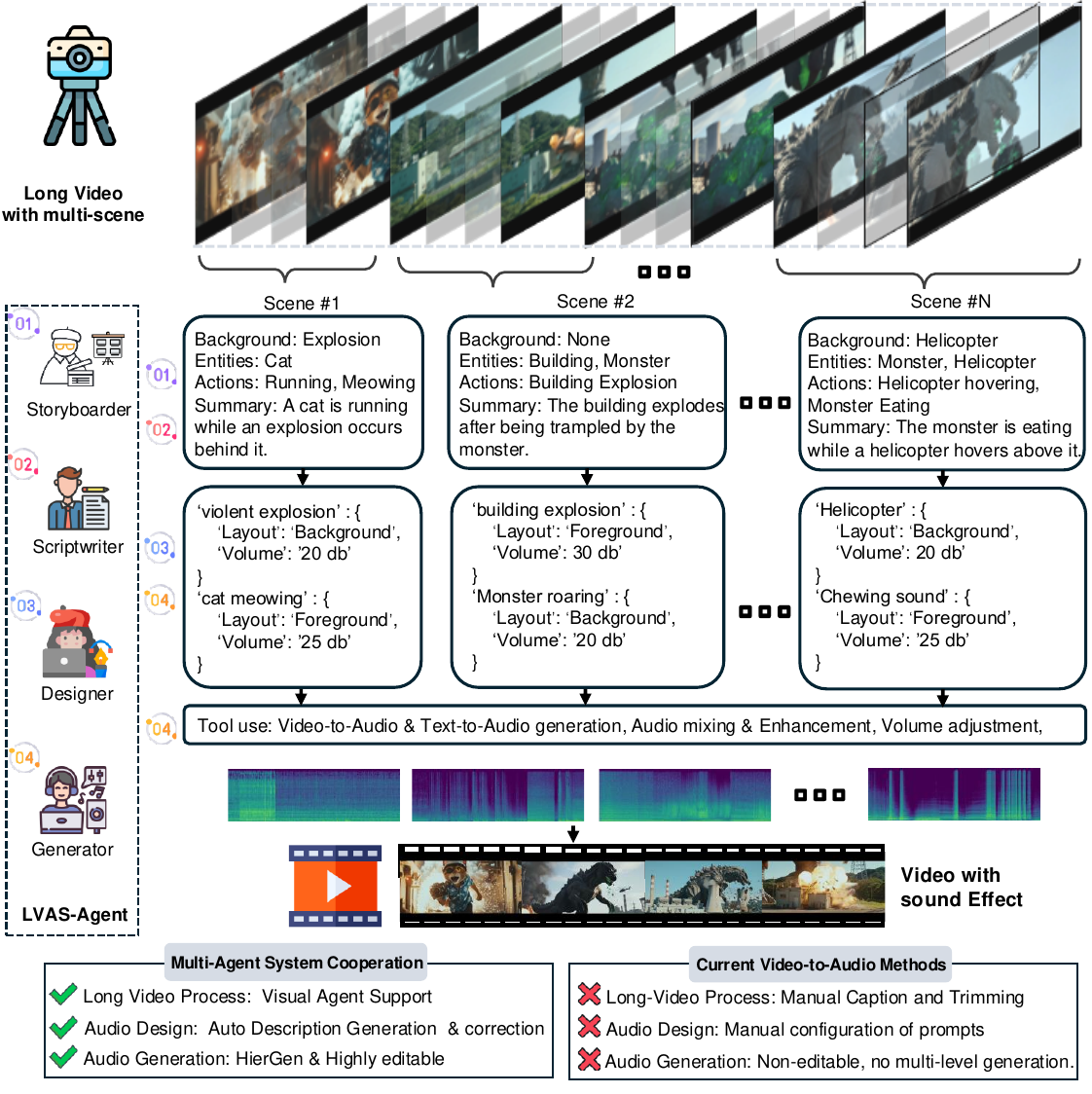}
    \captionsetup{font=small}
    \captionof{figure}{\textbf{We introduce LVAS-Agent, a multi-agent collaborative framework for end-to-end long video audio synthesis.} Built on VLM and LLM-based agents, it simulates real-world dubbing workflows, enabling automatic video script generation, audio design, and high-quality audio synthesis for long videos.}
    \label{fig:teaser}
\end{strip}

\begin{abstract}
Video-to-audio synthesis, which generates synchronized audio for visual content, critically enhances viewer immersion and narrative coherence in film and interactive media. However, video-to-audio dubbing for long-form content remains an unsolved challenge due to dynamic semantic shifts, temporal misalignment, and the absence of dedicated datasets. While existing methods excel in short videos, they falter in long scenarios (e.g., movies) due to fragmented synthesis and inadequate cross-scene consistency. We propose \textbf{LVAS-Agent}, a novel multi-agent framework that emulates professional dubbing workflows through collaborative role specialization. Our approach decomposes long-video synthesis into four steps including scene segmentation, script generation, sound design and audio synthesis. Central innovations include a discussion-correction mechanism for scene/script refinement and a generation-retrieval loop for temporal-semantic alignment. To enable systematic evaluation, we introduce \textbf{LVAS-Bench}, the first benchmark with 207 professionally curated long videos spanning diverse scenarios. Experiments demonstrate superior audio-visual alignment over baseline methods. Project page: \protect\phantomsection \href{https://lvas-agent.github.io}{https://lvas-agent.github.io}.
\end{abstract}

\section{Introduction}
\label{sec:intro}

Recent advancements in generative AI, particularly diffusion models and large language models (LLMs), have significantly improved short-video dubbing, enabling synchronized audio that enhances viewer immersion. However, long-video dubbing presents unique challenges, including complex semantic shifts, cross-scene temporal alignment, and adaptation to dynamic content. Current models, optimized for short clips, struggle to maintain narrative coherence over extended durations, limiting their effectiveness in applications such as film dubbing, AIGC video voiceovers, and automatic narration for mute videos. Moreover, the lack of dedicated datasets for long-video audio synthesis has hindered progress, as current datasets and benchmarks focus on short-form content.



Existing video-to-audio methods fall into two categories: (1) training dedicated generators such as SpecVQGAN~\cite{iashin2021taming} and Diff-Audio~\cite{popov2023optimal}, which capture short-term correlations but fail in long-term scene transitions, and (2) adapting text-to-audio models like SonicVisionLM~\cite{xie2024sonicvisionlm} and V2A-Mapper~\cite{wang2024v2a}, which heavily rely on textual descriptions and struggle with implicit visual cues in long videos. These methods encounter common issues:(i) they lack mechanisms to capture long-range dependencies across dynamically changing scenes, (ii) they fail to preserve contextual continuity in dialogue-rich videos, and (iii) they struggle to synthesize background sounds that evolve naturally over extended durations. Additionally, these methods rely on short-video datasets, which lack annotations for multi-sounds and cross-scene consistency with only 2-4 words for each audio labels.


A fundamental question arises: \textit{How can we leverage short-video dubbing priors to enable long-video synthesis while ensuring semantic coherence, temporal alignment, and scalable synthesis without requiring large-scale long-video training data?}
A naive approach is to split long videos into shorter segments and apply existing methods.  However, this approach practically may lead to issues such as poor continuity, unnatural transitions, and unclear main voice due to the lack of understanding of long-sequence videos.

To address these challenges, we present \textbf{LVAS-Agent}, a multi-agent framework that mimics professional dubbing workflows through structured role collaboration. Our key innovation lies in decomposing the synthesis process into specialized stages with collaborative agents: semantic-aware scene segmentation, context-sensitive script generation, ambiguity-resolved sound design, and knowledge-enhanced audio synthesis. The overall is shown in Figure~\ref{fig:teaser}.

Specifically, our method operates through four tightly coupled roles. The \textbf{Storyboarder} first segments videos into narrative-preserving scenes using shot transition detection and contrastive keyframe clustering. The \textbf{Scriptwriter} then generates time-aligned audio scripts by fusing visual semantics from CLIP-encoded~\cite{radford2021learning} features with dialogue context analysis. Building on this, the \textbf{Designer} employs spectral saliency analysis to disentangle foreground dialogues from ambient sounds, refining annotations through agent-mediated ambiguity resolution. Finally, the \textbf{Synthesizer} orchestrates hybrid audio generation, blending neural text-to-speech with diffusion-based environmental effects. Central to the system are two collaborative mechanisms: a discussion-correction process for scene merging and script refinement, and a generation-retrieval-optimization loop that iteratively aligns sound design with retrievable audio knowledge, enabling precise temporal and semantic consistency.

To enable systematic evaluation, we introduce the Long-Video Audio Synthesis (LVAS) benchmark, comprising 207 professionally selected long videos. \textbf{LVAS-Bench} covers various scenes such as urban landscapes, combat simulations, and animation actions to ensure the accuracy of the evaluation.

Our contributions can be summarized as follows:
\begin{itemize}
    \item We introduce LVAS-Agent, a multi-agent framework that systematically addresses long-video dubbing challenges by structuring the synthesis process into role-specialized collaborative agents.

    \item We release LVAS-Bench, the first dedicated long-video audio synthesis dataset, covering 207 professionally curated videos across diverse scenarios, enabling standardized benchmarking.
    \item Experiments demonstrate that LVAS-Agent improves semantic alignment, temporal alignment and distribution matching of audio-visual for long-video dubbing over existing baselines.
\end{itemize}

\section{Related Work}
\label{sec:related}

\subsection{Video-to-Audio Generation}
Video-to-audio generation, also known as dubbing, is a crucial audio technique for enhancing viewers' auditory experience and has witnessed significant evolution with the advent of neural approaches. Early neural dubbing models demonstrated deep learning's potential in sound effect creation, though limited to specific genres \cite{chen2018visually, chen2020generating, mo2024text, zhou2018visual}. Recent advancements in video-to-audio generation have followed two main directions. The first approach focuses on training generators from scratch, with notable works including SpecVQGAN \cite{iashin2021taming}, which employs a cross-modal Transformer for auto-regressive sound generation, Im2Wav \cite{sheffer2023hear}, which conditions audio generation on CLIP features, Diff-Foley \cite{luo2023diff}, which enhances alignment through contrastive pre-training and MMAudio \cite{cheng2024taming}, which use a flow matching-based multimodal joint training framework on large-scale data. The second approach adapts text-to-audio generators, with Xing et al. \cite{xing2024seeing} utilizing ImageBind \cite{girdhar2023imagebind} for optimization-based alignment, SonicVisionLM \cite{xie2024sonicvisionlm} employing caption-based synthesis, V2A-Mapper \cite{wang2024v2a} directly translating visual to text embeddings and FoleyCrafter \cite{zhang2024foleycrafter} integrating a learnable module into text-to-audio models with end-to-end training. Despite these advances, existing methods primarily excel only with short videos, encountering issues such as noise artifacts and audio-scene inconsistencies in longer videos. Our method addresses these limitations by incorporating a video understanding segmentation module, providing an effective solution for audio generation in long-form videos.

\subsection{MLLMs for Video Understanding}
Recent advances in vision foundation models \cite{dosovitskiy2020image,liu2021swin,radford2021learning,tianIntegrallyPreTrainedTransformer2023,tian2021semantic,zhang2022hivit,kirillovSegmentAnything2023} have led to the emergence of multimodal LLMs (MLLMs) \cite{liuVisualInstructionTuning2023,tian2024chatterbox,zhangGPT4RoIInstructionTuning2023,laiLISAReasoningSegmentation2023}, which have demonstrated remarkable capabilities in language-guided visual understanding. This progress has naturally extended to video understanding, with pioneering works including VideoChat \cite{li2024videochat}, Video-ChatGPT \cite{maaz2023videochatgpt}, Video-LLaMA \cite{zhang2023videollama}, Video-LLaVA \cite{lin2023videollava}, LanguageBind \cite{zhu2024languagebind}, and Valley \cite{luo2023valley}. However, videos present unique challenges compared to static images, particularly due to their temporal nature and the substantial volume of visual information that, when tokenized, often exceeds MLLMs' context limitations. While most existing approaches address this through frame sampling, some methods, such as Video-ChatGPT \cite{maaz2023videochatgpt}, have introduced more efficient video feature representations. The field has also witnessed significant progress in instance-level video understanding, with works like PG-Video-LLaVA \cite{munasinghe2023pgvideollava} for video grounding, and Artemis \cite{qiu2024artemis} for video referring, expanding the capabilities of video understanding systems.

The challenge becomes particularly acute in long video understanding, where effective keyframe selection becomes more crucial and complex. While some approaches like Kangaroo \cite{liu2024kangaroo} and LLaVA-Video \cite{zhang2024video} leverage language models with expanded context windows to accommodate more frames, others have developed specialized techniques to handle this limitation. MovieChat \cite{song2024moviechat}, for instance, implements a dual-memory system with short-term and long-term memory banks for efficient video content compression and preservation. Similarly, MA-LMM \cite{he2024ma} and VideoStreaming \cite{qian2024streaming} employ a Q-former alongside a compact language model (Phi-2 \cite{javaheripi2023phi}) for video data condensation. LongVLM \cite{weng2024longvlm} takes a different approach by utilizing token merging to reduce video token count.

\subsection{Multi-Agent System}
Multi-agent systems have evolved significantly with the advent of large language models (LLMs). Early single-agent frameworks like ModelScope-Agent \cite{li2023modelscope} and Toolformer \cite{schick2023toolformer} demonstrated the potential of LLMs in executing complex tasks through tool integration. HuggingGPT \cite{shen2023hugginggpt} and AudioGPT \cite{huang2024audiogpt} further expanded this capability by incorporating domain-specific models and functionalities.

However, the limitations of single-agent systems led to the emergence of multi-agent frameworks. Inspired by the Society of Mind \cite{minsky1988society}, works like Generative Agents \cite{park2023generative} pioneered the development of ``Simulated Society", where multiple agents interact within defined environments. Practical implementations such as ChatDev \cite{qian2023chatdev}, MetaGPT \cite{hong2023metagpt}, and TransAgents \cite{wu2024transagents} have successfully demonstrated collaborative problem-solving through simulated workflows, achieving superior reasoning and factuality compared to single-agent approaches. These systems effectively address complex tasks requiring meaningful collaborative interaction, which single agents typically struggle to accomplish.

For long-video audio synthesis, while AudioAgent \cite{wang2024audio} employs pre-trained diffusion models and GPT-4, it lacks explicit multi-agent collaboration and specific optimizations for long videos. In contrast, we propose LVAS-Agent: a multi-agent collaborative framework that mimics professional dubbing workflows. Our system comprises specialized agents for video segmentation, content understanding (leveraging advanced MLLM models for precise description), and audio tag generation (separating foreground and background elements). These components work in concert with MMAudio to produce high-quality synthesized audio.

\section{Method}
\subsection{Overview}

\begin{figure*}
  \centering
\includegraphics[width=1\linewidth]{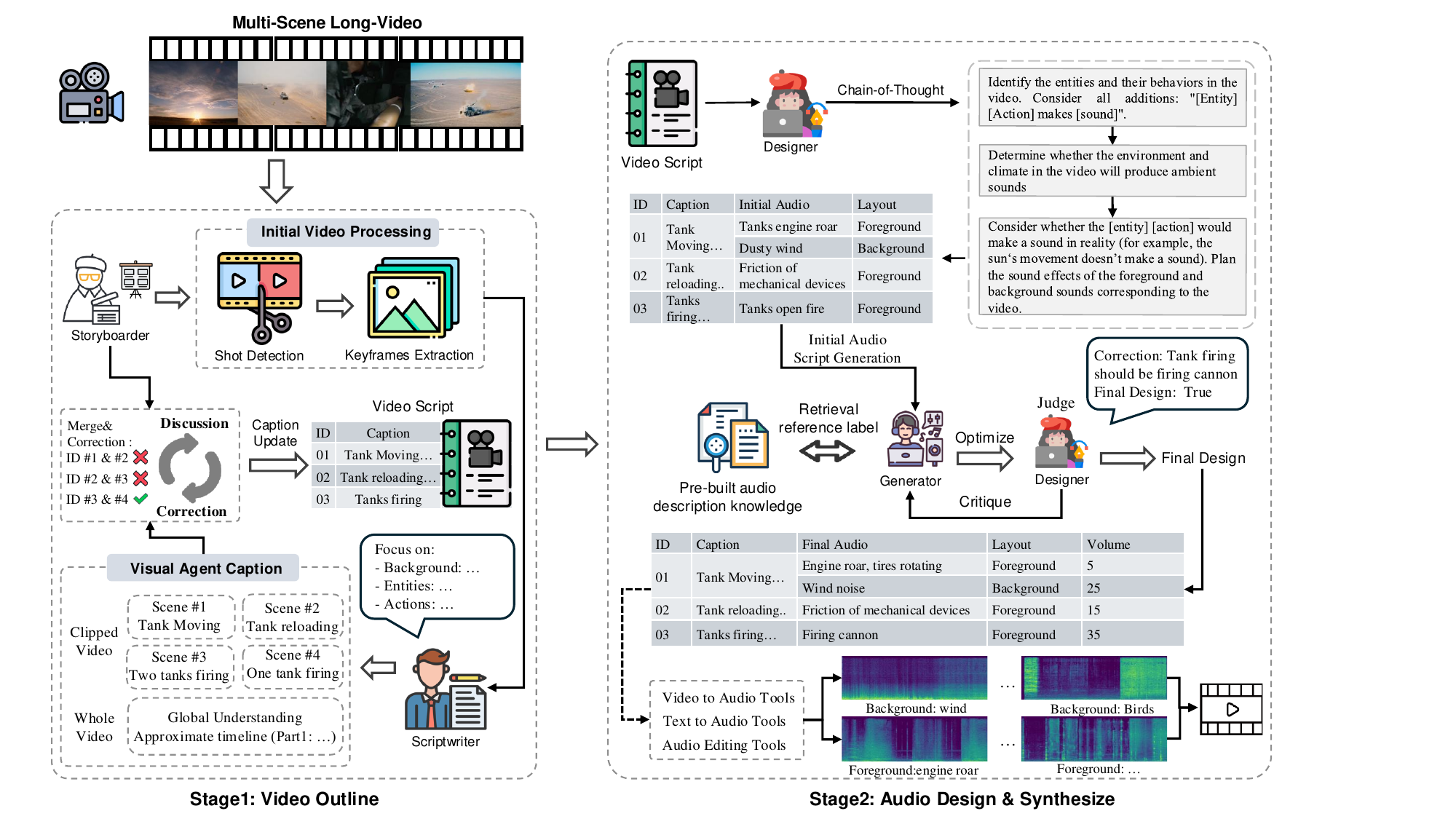}
   \vspace{-4mm}
   \caption{\textbf{Workflow of LVAS-Agent}. Given the original video, Storyboarder and Scriptwriter collaborate through Discussion and Correction to create a structured video script. The Designer and Generator complete multi-layered, high-quality sound synthesis through the Generate-Retrieve-Optimize mechanism.}
   \vspace{-4mm}
   \label{fig:framework}
\end{figure*}

By clearly defining the roles of agents, multi-agent systems can decompose complex tasks into smaller, more manageable ones. In LVAS-Agent, we define four main characters: \textbf{Storyboarder}, \textbf{Scriptwriter}, \textbf{Designer}, and \textbf{Synthesizer}. As show in Figure~\ref{fig:framework}, each of these roles carries its own specific set of responsibilities. The \textbf{Storyboarder} is responsible for the creation of video storyboards. This includes planning the storyboard strategy, segmenting video scenes, and extracting keyframes. The \textbf{Scriptwriter} is in charge of writing the video script. Their responsibilities include understanding the video content, collaborating with the storyboard artist to generate a detailed video outline, and providing references for the sound designer. The \textbf{Designer} is tasked with annotating sound effects based on the video outline. This includes analyzing video descriptions, generating detailed sound effect annotations for each potential sound, classifying entity and environmental sounds, and collaborating with the voiceover artist to refine the sound annotations. Finally, the \textbf{Generator} is responsible for the actual sound effect synthesis. This includes transforming sound effect annotations into suitable sound labels and using professional tools to achieve step-by-step sound-video synthesis, combining the main sound and background sound.

\subsection{Multi-agent collaboration strategy}
In this section, we present the two agent collaboration strategies employed in this work: \textit{Discussion-Correction} (Algorithm \ref{alg:1}) and \textit{Generation-Retrieval-Optimization} (Algorithm \ref{alg:2}).

\noindent\textbf{Discussion-Correction}
This strategy, as illustrated by Algorithm \ref{alg:1}, is executed through the collaboration between two agents . First, the \textit{Storyboarder agent} \( \mathbf{P} \) segments the video into distinct scenes, denoted as \([v_0, \dots, v_n]\), based on shot transitions and extracts the corresponding keyframes lists  \( \{[kf_{1,1}, \dots], \dots, [kf_{n,1}, \dots ] \}\). Next, the \textit{Scriptwriter agent} \( \mathbf{Q} \) performs a global analysis of the entire video, followed by a detailed examination of each segment based on its respective keyframes. The Storyboarder agent \(\mathbf{P}\) and Scriptwriter agent \(\mathbf{Q} \) then engage in a discussion to determine whether certain segments should be merged and whether the segment captions require refinement, based on both the global understanding and the detailed video captions. The final output is a structured video script.

\begin{algorithm}[htbp]
\caption{Collaboration Strategy}\label{alg:1}

\KwIn{Storyboarder agent \(\mathbf{P}\), Scriptwriter agent \(\mathbf{Q}\), Video \(\mathbf{V}\)}
\KwOut{Structured video script \(\mathbf{T}\)}

\SetKwData{T}{$\mathbf{T}$}
\SetKwData{V}{$\mathbf{V}$}
\SetKwData{P}{$\mathbf{P}$}
\SetKwData{Q}{$\mathbf{Q}$}
\SetKwData{U}{$\mathrm{U}$}
\SetKwData{D}{$D$}
\SetKwData{Merge}{$\text{merge}$}

\T{} $\gets \emptyset$ 
$\{v_0, \dots, v_n\} \gets \P(\V)$ \tcp{Shot Change Detection}
$\{kf_0, \dots, kf_n\} \gets \P(\{v_0, \dots, v_n\})$  \tcp{Keyframe Extraction}
$\U_{\V} \gets \Q(\V)$ \tcp{Understand global content, style features} 
\For{$i = 0$ \KwTo $n$}{
    $\U_i \gets \Q(kf_i)$  \\
    \If{$i > 1$}{
        $D \gets \P(\U_i, \U_{i-1}, \U_{\V})$ \\
        \If{$D = MERGE$}{
            $v_{i-1} \gets merge\_segments(v_{i-1}, v_i)$ 
            $\T \gets \T \cup [\U_{i-1}, \U_i]$ 
        }
        \Else{
            $\T \gets \T \cup [\U_i]$ 
        }
    }
}
$\T \gets \T \cup \U_{\V}$  \\
\KwRet{$\T$}

\end{algorithm}

\noindent\textbf{Generation-Retrieval-Optimization} This is accomplished through the collaboration between the \textit{Designer agent} \(\mathbf{D}\) and the \textit{Synthesizer agent} \(\mathbf{S}\). First, the Designer agent \(\mathbf{D}\) formulates an initial sound design based on the video script. The Synthesizer agent \(\mathbf{S}\) then retrieves relevant knowledge from a sound synthesis database to generate a concrete implementation plan. This plan is reviewed by the Designer agent \(\mathbf{D}\), who decides whether further refinement of the sound design is needed or if the plan is ready for final synthesis. Specifically, this process begins with an in-depth understanding of the video script, followed by iterative exchanges of feedback between the Designer agent \(\mathbf{D}\) and the Synthesizer agent \(\mathbf{S}\). Through multiple iterations, the final sound synthesis plan is determined.

\begin{algorithm}[htbp]
\caption{Generation-Retrieval-Optimization Collaboration Strategy}
\label{alg:2}
\SetAlgoNlRelativeSize{0}
\KwIn{Designer agent \(\mathbf{D}\), Synthesizer agent \(\mathbf{S}\), Video script \(\mathbf{T}\), Maximum iterations \(N_{\max}\)}
\KwOut{Finalized sound synthesis plan \(\mathbf{A}_{\text{final}}\)}

\textbf{Initialization:} \( \mathbf{A}_{\text{init}} \gets \mathbf{D}(\mathbf{T}) \); \\
\( \mathbf{A}_{\text{retrieved}} \gets \mathbf{S}(\mathbf{A}_{\text{init}}) \);

\For{\( i = 1 \) to \( N_{\max} \)}{
    
    \( \mathbf{A}_{\text{reviewed}} \gets \mathbf{D}(\mathbf{A}_{\text{retrieved}}) \); \\
    \If{\(\mathbf{D}\) determines \(\mathbf{A}_{\text{reviewed}}\) is FINAL}{
        \textbf{break}; \tcp{Exit early if finalized}
    }
    \( \mathbf{A}_{\text{modified}} \gets \mathbf{D}(\mathbf{A}_{\text{reviewed}}) \); \\
    \( \mathbf{A}_{\text{retrieved}} \gets \mathbf{S}(\mathbf{A}_{\text{modified}}) \);
}

\Return \( \mathbf{A}_{\text{reviewed}} \);

\end{algorithm}

\subsection{Video Structure}
As shown in Figure~\ref{fig:framework}, this paper proposes a structured video script generation method to assist in generating sound effects for full-length videos. The method addresses three core challenges: 1) existing audio synthesis tools have duration constraints; 2) current Video-To-Audio methods struggle with scene and content transitions, hindering semantic and temporal alignment; 3) ensuring consistency between video captions and audio descriptions for coherent synthesis. To overcome these, we introduce a fine-grained video structuring approach, supported by collaboration between storyboarder and scriptwriter agents, as outlined in Algorithm \ref{alg:1} The specific design of these agents is detailed as follows.

\noindent\textbf{Storyboarder} is an LLM-based agent responsible for fine-grained video structuring in the VTA task. Its key functions include detecting shot transitions for coarse segmentation, extracting key frames using the K-Means clustering algorithm, and refining segment boundaries and captions based on the Scriptwriter's Video Caption. Shot detection uses an HSV color space transition method for rapid, frame-accurate segmentation, enabling detailed video understanding. By extracting key frames from smaller segments, it captures more visual information compared to directly inputting the full video into a vision-language model, enhancing video comprehension. Storyboarder also collaborates with the Scriptwriter to decide whether segments should be merged or captions refined, considering both local and global video context. Detailed implementation is in Appendix \textbf{\ref{fig:Storyboarder}}.

\noindent\textbf{Scriptwriter} is a visual support agent responsible for comprehending both the full video and individual video segments. Recent video understanding tasks achieve comprehension by extracting information from visual contexts to derive semantic features \cite{li2024videochatchatcentricvideounderstanding} or by directly generating descriptive text \cite{fan2025videoagent}. Textual descriptions of the video script make it easier to maintain consistency between video and audio descriptions. Furthermore, the textual format facilitates interaction among multiple agents. Notably, transforming the video into a structured script, independent of video frames, enhances processing speed and significantly reduces the number of tokens. The detailed implementation is provided in Appendix \ref{fig:Scriptwriter_full}.

\begin{figure*}[htbp]
  \centering
   \includegraphics[width=1\linewidth]{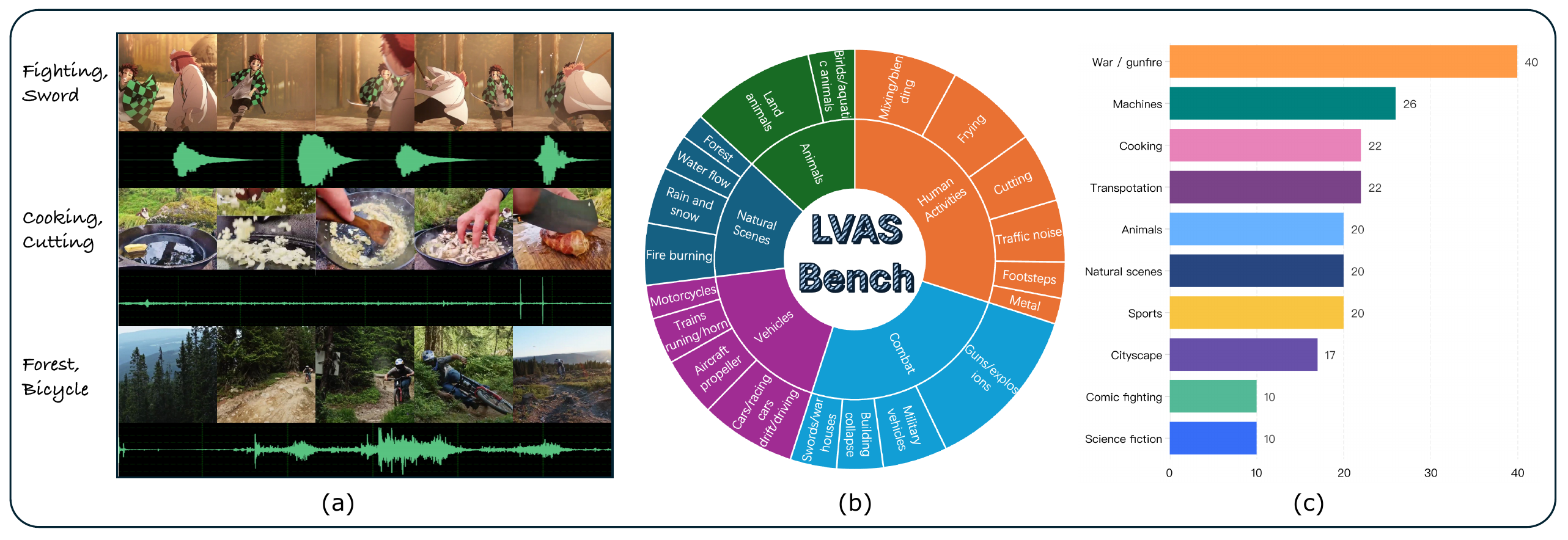}
   \vspace{-4mm}
   \caption{Our LVAS-Bench is presented in the following parts: (a) illustrates sample data from the benchmark, (b) provides statistical distributions of audio categories and sub-categories across the dataset, and (c) presents the statistics of video categories within the dataset.}
   \vspace{-4mm}
   \label{fig:benchmark}
\end{figure*}

\subsection{Audio Design and Generation}
This section presents the second stage of LVAS-Agent: audio design and generation, as shown in Figure~\ref{fig:framework}. The design follows key principles: 1) Mimicking professional sound design workflows by analyzing video scripts for accurate audio descriptions, 2) Enhancing efficiency and quality using existing audio generation tools, and 3) Ensuring structured, editable audio planning for fine-grained control. This stage uses a collaborative framework with two LLM-based agents, integrating retrieval-augmented generation (RAG) and audio synthesis tools to create high-quality, multi-layered audio.

\noindent\textbf{Designer} annotates audio in the video script and collaborates with the Synthesizer agent to finalize the audio design. Real-world dubbing often involves complex scenes with layered environmental sounds, diverse sound-producing actions, and varying audio levels. To address this, we introduce a Chain-of-Thought (CoT) reasoning mechanism, breaking the task into steps: identifying primary action sounds, analyzing background audio, and ensuring audio coherence. The Designer agent creates the initial audio design, covering foreground and background sounds, volume control, and sound descriptions, while verifying alignment with the video content. It then provides iterative feedback to the Audio Synthesizer to optimize the final audio plan.

\noindent\textbf{Generator} 
The Generator synthesizes audio based on the audio annotations obtained through collaboration with the Designer. It uses retrieval-augmented generation (RAG) with an audio label knowledge base, Video-to-Audio (VTA) and Text-to-Audio (TTA) models for synthesis, hierarchical mixing, and volume adjustment. RAG-based retrieval ensures high-quality synthesis, addressing the limitations of VTA models trained on the VGGSound dataset, which contains only 310 audio labels with 2-4 words each. When audio prompts match these predefined labels, the generated audio is more stable and higher quality.

Building on this insight, all VGGSound labels were reorganized and reclassified into 20 common video scenarios. To enrich the labels, GPT-4 and human annotators added details such as typical scenarios and relevant objects or interactions. This resulted in 192 refined labels. The structured knowledge base allows the Generator to retrieve and modify predefined labels, rather than relying on open-ended prompts. This retrieval-based approach enables the “generation-retrieval-optimization” mechanism in Algorithm \ref{alg:2}, facilitating iterative refinement of audio synthesis. The LVAS-Agent employs MMAudio \cite{cheng2024taming}, an open-source framework supporting VTA and TTA synthesis, ensuring flexibility for final audio mixing, volume adjustment, and refinements.

\section{LVAS-Bench}
\textbf{Collection.} We construct the first specialized \textbf{l}ong-\textbf{v}ideo \textbf{a}udio \textbf{s}ynthesis \textbf{bench}mark(\textbf{LVAS-Bench}). The benchmark contains 207 professionally curated videos (with an average duration of 1 minute) sourced from three main origins: (1) film production archives with open licenses, (2) annotated documentary segments, and (3) procedurally generated synthetic scenes. Importantly, all videos in the dataset have pure sound effects, with no background noise or human speech. This creates a dataset of long-form, semantically rich videos with clear transitions, matched with corresponding pure sound-audio. Figure~\ref{fig:benchmark}(a) illustrates representative video-audio cases.

\begin{figure*}[htbp]
  \centering
\includegraphics[width=0.85\linewidth]{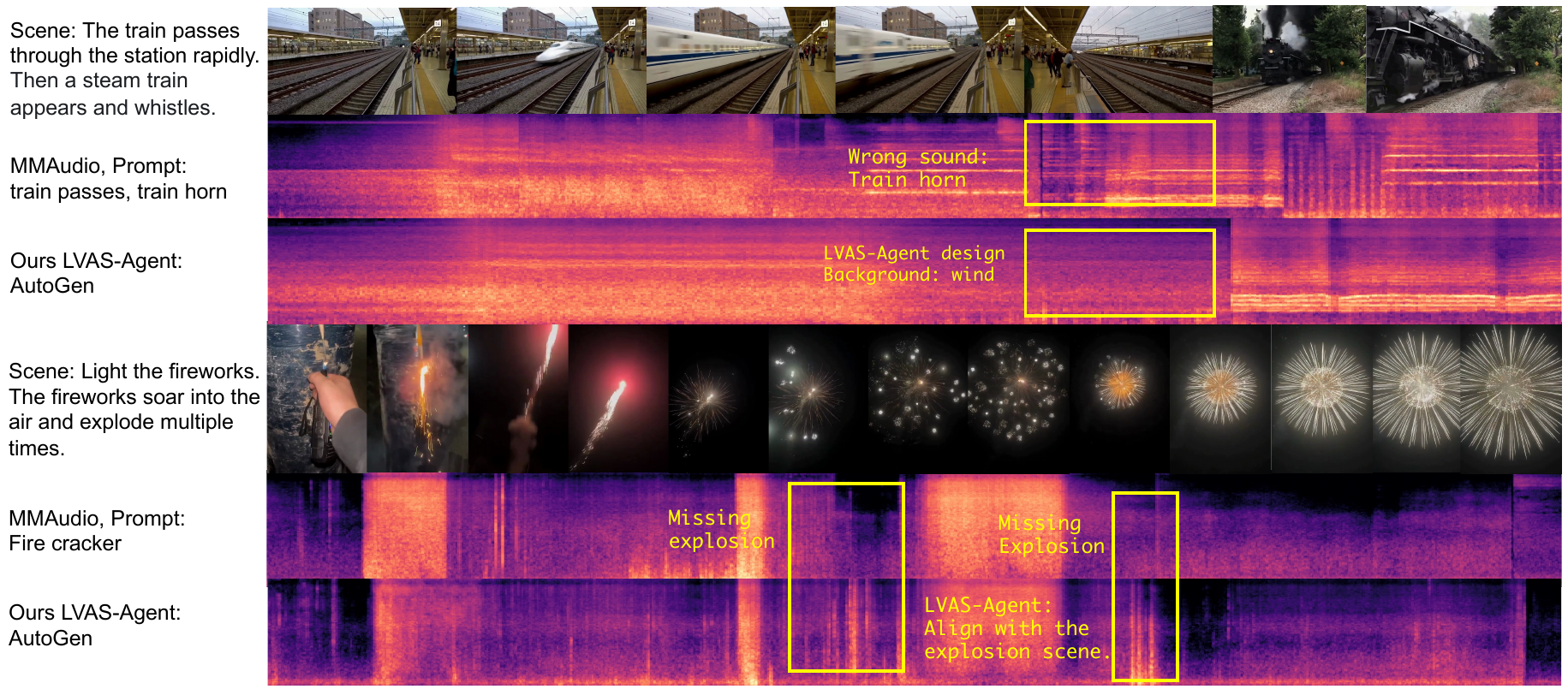}
   \vspace{-4mm}
   \caption{We visualize the spectrograms of generated audio (by prior works and our method). LVAS-Agent demonstrates superior performance in synthesizing long video audio, ensuring seamless scene transitions without errors or missing sounds.}
   \vspace{-4mm}
   \label{fig:visual}
\end{figure*}

\noindent\textbf{Statistical Analysis.} To ensure diversity, LVAS-Bench covers sufficient video and audio categories. Figure~\ref{fig:benchmark}(b) visualizes the benchmark's audio types, encompassing five major classes (e.g., human activities) with numerous fine-grained subcategories. Figure~\ref{fig:benchmark}(c) quantifies the distribution across 10 video-level categories, where instances such as the ``cooking" category comprise 22 entries.

\noindent\textbf{Benchmark Annotation.} LVAS-Bench also offers detailed time-stamped annotations for each video-audio pair and comprehensive global descriptions. The time-stamped annotations indicate captions from specific seconds to specific seconds, while the global descriptions provide a detailed account of the entire long video. We implement a hybrid annotation protocol: initial annotations are generated by video understanding model, subsequently refined through manual verification by domain experts.

\definecolor{lightpurple}{RGB}{235, 222, 240}
\definecolor{lightblue}{RGB}{200, 230, 255}   
\definecolor{lightgreen}{RGB}{200, 255, 200}  
\definecolor{lightorange}{RGB}{255, 200, 180}

\section{Experiment}
\begin{table*}[t]
    \centering
    \renewcommand{\arraystretch}{1.2} 
    \setlength{\tabcolsep}{0.8pt} 
    \small
    \begin{tabular}{c|>{\columncolor{lightpurple}}c>{\columncolor{lightpurple}}c>{\columncolor{lightpurple}}c>{\columncolor{lightpurple}}c>{\columncolor{lightpurple}}c|>{\columncolor{lightblue}}c>{\columncolor{lightblue}}c|>{\columncolor{lightgreen}}c|>{\columncolor{orange!50}}c}
        \toprule
        \multirow{2}{*}{\textbf{Methods}} & \multicolumn{5}{c|}{\cellcolor{lightpurple}\textbf{Distribution Matching}} & \multicolumn{2}{c|}{\cellcolor{lightblue}\textbf{Audio Quality}} & \cellcolor{lightgreen}\textbf{Semantic Align} & \cellcolor{orange!50}\textbf{Temporal Align} \\
        \cline{2-10}
        & \(\mathrm{FD_{VGG}}\)$\downarrow$ & \(\mathrm{FD_{PANN}}\)$\downarrow$ & \(\mathrm{FD_{PASST}}\)$\downarrow$ & \(\mathrm{KL_{PANNs}}\)$\downarrow$ & \(\mathrm{KL_{PASST}}\)$\downarrow$ & \(\mathrm{IS_{PANNs}}\)$\uparrow$ & \(\mathrm{IS_{PASST}}\)$\uparrow$ & IB-Score$\uparrow$ & DeSync$\downarrow$ \\
        \hline \hline
        Baseline (FoleyCrafter) & 6.61 & 60.66 & 637.82 & 2.68 & 2.65 & 4.79 & 4.34 & 0.28 & 1.24 \\
        Baseline (MMAudio) & 9.48 & 51.73 & 588.24 & 2.02 & 1.80 & 3.91 & 3.05 & 0.32 & 0.61 \\
        \textbf{LVAS-Agent (Ours)} & \textbf{5.76} & \textbf{46.16} & \textbf{573.67} & \textbf{1.86} & \textbf{1.77} & 4.28 & 3.50 & \textbf{0.33} & \textbf{0.53} \\
        \bottomrule
    \end{tabular}
    \caption{Comparison of different methods on various evaluation metrics. Lower values ($\downarrow$) indicate better performance, while higher values ($\uparrow$) indicate better quality.}
    \label{tab:comparison}
\end{table*}

\begin{table*}[h]
    \renewcommand{\arraystretch}{1.2}
    \small
    \centering
    \begin{tabular}{ccccccccc}
        \toprule
        \multicolumn{3}{c}{Key Components} & \multicolumn{2}{c}{Distribution Matching} & \multicolumn{1}{c}{Audio Quality} & \multicolumn{1}{c}{Semantic Align} & \multicolumn{1}{c}{Temporal Align} \\
        \cmidrule(lr){1-3} \cmidrule(lr){4-5} \cmidrule(lr){6-6} \cmidrule(lr){7-7} \cmidrule(lr){8-8}
        Video-Structure & Chain-of-Thought & RAG & \(\mathrm{FD_{VGG}}\)$\downarrow$ & \(\mathrm{FD_{PANNs}}\)$\downarrow$ & \(\mathrm{IS_{PNSS}}\)$\downarrow$ & IB-Score$\uparrow$ & DeSync$\uparrow$ \\
        \hline \hline
        \checkmark &  &  & 7.45 & 77.65 & 1.85 & 0.312 & 0.361 \\
        \checkmark & \checkmark &  & 7.41 & 76.84 & 1.82 & 0.319 & 0.346 \\
        \checkmark & \checkmark &\checkmark & 7.12 & 71.61 & 1.81 & 0.336 & 0.338 \\
        \bottomrule
    \end{tabular}
    \caption{\textbf{Ablation Study.} Ablating different key components of LVAS-Agent and evaluating performance on LVAS-Bench.}
    \label{tab:ablation}
\end{table*}

\subsection{Experiment Setup.} 
\textbf{Metrics} We assess the generation quality in four different dimensions: distribution matching, audio quality, semantic alignment, and temporal alignment. 1) Distribution matching assesses the similarity in feature distribution between ground-truth audio and generated audio, under some embedding models. We compute Fréchet Distance (FD) and Kullback–Leibler (KL) distance. For FD, we adopt PaSST \cite{koutini22passt} ((\(\mathrm{FD_{PaSST}}\)), PANNs \cite{kong2020panns} (\(\mathrm{FD_{PANNs}}\)), and VGGish \cite{vggish}  as embedding models. For the KL distance, we adopt PANNs (\(\mathrm{KL_{PANNs}}\)) and PaSST (\(\mathrm{KL_{PaSST}}\)) as classifiers. 2) We use PANNs as the classifier, following Wang et al. \cite{wang2024frieren}, to assess generation audio quality without the need for comparison with the ground truth, utilizing the inception score. 3) Semantic alignment is measured using ImageBind \cite{10203733}, following Viertola et al. \cite{viertola2024temporallyalignedaudiovideo}, by extracting visual features from the input video and audio features from the generated audio, then computing the average cosine similarity as the IB-score. 4) Temporal alignment: We use synchronization score (DeSync) to assess audio-visual synchrony. DeSync is predicted by Synchformer \cite{huang2023makeanaudio2temporalenhancedtexttoaudio} as the misalignment (in seconds) between the audio and video.

\noindent\textbf{Data.} Since our method focuses on the task of sound effect synthesis for long videos, which consist of shorter video-audio pairs, are not suitable for evaluation. Therefore, this paper uses the proposed LVAS-Bench to assess the performance of the Agent-System.

\noindent\textbf{Baselines.} To accommodate sound effect synthesis for videos of arbitrary length, the experimental baseline is designed to first segment the video, then apply Video-to-Text (VTA) on each segment, and finally combine the results. We use state-of-the-art open-source methods, FoleyCrafter \cite{zhang2024foleycrafter} and MMAudio \cite{cheng2024taming}, as VTA tools. FoleyCrafter supports audio generation for segments up to 10 seconds, while MMAudio performs better for videos around 10 seconds due to the duration of its training data. Consequently, we set the segment interval for the baseline to 10 seconds.

\noindent\textbf{Implementation Details.} In our experiments, all LLM-based agents use the Qwen API \cite{qwen2025qwen25technicalreport} with the ``qwen-max" model to simulate different agent roles. The visual support agent is implemented using the locally deployed ``Qwen2.5-VL-7B" model. The retrieval-augmented generation for the predefined audio description knowledge base is built on LlamaIndex \footnote{LlamaIndex: \href{https://www.llamaindex.ai/}{https://www.llamaindex.ai/}.} and powered by a ``qwen-plus" model.

\subsection{Main Results}
The evaluation metric comparison results are shown in Table \ref{tab:comparison}, where LVAS-Agent outperforms the baseline methods across all metrics in four key dimensions, achieving state-of-the-art performance. Additionally, we visualize and compare the audio waveforms generated by different methods. The quantitative results demonstrate that our approach enables the existing VTA base models to generate higher-quality audio in long videos with enhanced semantic and temporal consistency, all without additional training. As shown in the visualized spectrogram comparison in Figure~\ref{fig:visual}, (a) reveals that LVAS-Agent exhibits adaptive capability to video content variations, ensuring a high level of alignment with the video while reducing the omission of key sound effects and minimizing incorrect audio generation. Furthermore, (b) shows that our method, by designing foreground and background audio layers, achieves a multi-level synthesis that enhances its off-screen capability.

\subsection{Ablation Study}
To validate the effectiveness of the Agent-Framework design, an ablation study was conducted, as shown in Table \ref{tab:ablation}. Key components of the LVAS-Agent that contribute to enhancing audio generation quality were identified, including: (1) generating sound effects after refined video segmentation, (2) a Chain-of-Thought process for structured sound effect description and hierarchical generation, and (3) an iterative optimization process leveraging retrieved audio reference documents. The experiment was conducted on 20 randomly selected video cases. 

First, integrating the proposed video structuring method into the baseline significantly improved audio quality. This improvement is attributed to content-aware segmentation, which ensures consistency between video content and generated audio. Building on this, incorporating the CoT process for audio description further enhanced both audio-visual synchronization and quality. This is due to CoT’s ability to effectively transform video captions into detailed audio descriptions, systematically reasoning through possible sound effects and accurately identifying appropriate sources. Finally, the retrieval-augmented iterative optimization of audio descriptions further refined the VTA tool’s audio generation, leveraging a domain-specific knowledge base to translate LLM-generated generalized descriptions into precise audio prompts familiar to the VTA model.

\subsection{User Study}
\begin{figure}[htbp]
  \centering
\includegraphics[width=1\linewidth]{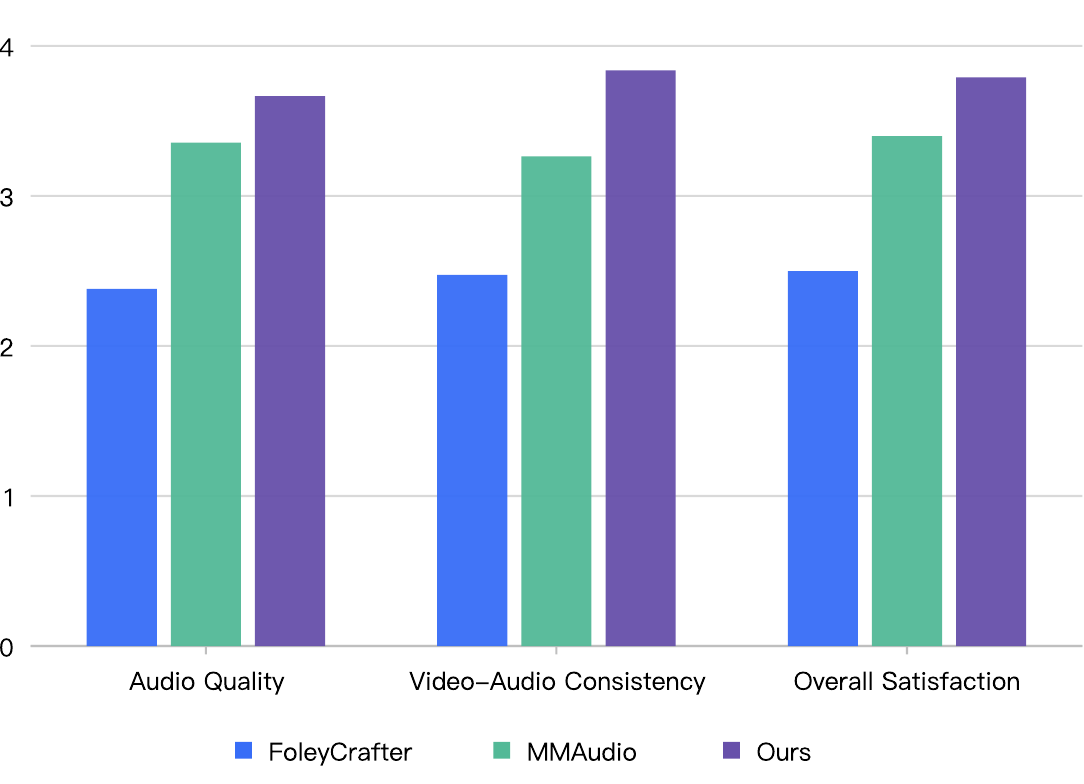}
   \caption{User study comparing our method with baselines across different aspects. Higher values indicate greater user preference.}
   \label{fig:user_study}
\end{figure}

We conducted a user study involving 30 participants to evaluate our method in comparison with FoleyCrafter \cite{zhang2024foleycrafter} and MMAudio \cite{cheng2024taming}. Participants were asked to listen to 10 audio samples generated by each method and rate them on a scale of 1 to 5 across three dimensions: “Audio Quality,” “Video-Audio Consistency,” and “Overall Satisfaction.” Higher scores indicate better performance. As illustrated in Figure~\ref{fig:user_study}, the results of the user study demonstrate that our method outperforms the two baseline approaches across all evaluated aspects.

\vspace{-5mm}
\section{Conclusion}
 We present LVAS-Agent, a multi-agent framework that systematically tackles long-video dubbing challenges through role-specialized collaborative agents. By decomposing the workflow into scene segmentation, script generation, sound design, and hybrid synthesis, our method overcomes limitations in semantic continuity and temporal alignment inherent to existing approaches. We also release the first dedicated long-video audio synthesis dataset, covering 207 professionally curated videos, named LVAS-Bench. Experimental results demonstrate superior performance in distribution matching, audio quality, and alignment metrics on LVAS-Bench.

For future work, we aim to develop a large-scale, finely annotated dataset of long-video audio to further advance the development of long-video dubbing models.
{
    \small
    \bibliographystyle{ieeenat_fullname}
    \bibliography{main}

\begin{thebibliography}{60}
\providecommand{\natexlab}[1]{#1}
\providecommand{\url}[1]{\texttt{#1}}
\expandafter\ifx\csname urlstyle\endcsname\relax
  \providecommand{\doi}[1]{doi: #1}\else
  \providecommand{\doi}{doi: \begingroup \urlstyle{rm}\Url}\fi

\bibitem[Chen et~al.(2018)Chen, Zhang, Fang, Wang, Bui, and Nevatia]{chen2018visually}
Kan Chen, Chuanxi Zhang, Chen Fang, Zhaowen Wang, Trung Bui, and Ram Nevatia.
\newblock Visually indicated sound generation by perceptually optimized classification.
\newblock In \emph{Proceedings of the European Conference on Computer Vision (ECCV) Workshops}, pages 0--0, 2018.

\bibitem[Chen et~al.(2020)Chen, Zhang, Tan, Xiao, Huang, and Gan]{chen2020generating}
Peihao Chen, Yang Zhang, Mingkui Tan, Hongdong Xiao, Deng Huang, and Chuang Gan.
\newblock Generating visually aligned sound from videos.
\newblock \emph{IEEE Transactions on Image Processing}, 29:\penalty0 8292--8302, 2020.

\bibitem[Cheng et~al.(2024)Cheng, Ishii, Hayakawa, Shibuya, Schwing, and Mitsufuji]{cheng2024taming}
Ho~Kei Cheng, Masato Ishii, Akio Hayakawa, Takashi Shibuya, Alexander Schwing, and Yuki Mitsufuji.
\newblock Taming multimodal joint training for high-quality video-to-audio synthesis.
\newblock \emph{arXiv preprint arXiv:2412.15322}, 2024.

\bibitem[Dosovitskiy et~al.(2020)Dosovitskiy, Beyer, Kolesnikov, Weissenborn, Zhai, Unterthiner, Dehghani, Minderer, Heigold, Gelly, et~al.]{dosovitskiy2020image}
Alexey Dosovitskiy, Lucas Beyer, Alexander Kolesnikov, Dirk Weissenborn, Xiaohua Zhai, Thomas Unterthiner, Mostafa Dehghani, Matthias Minderer, Georg Heigold, Sylvain Gelly, et~al.
\newblock An image is worth 16x16 words: Transformers for image recognition at scale.
\newblock \emph{arXiv preprint arXiv:2010.11929}, 2020.

\bibitem[Fan et~al.(2025)Fan, Ma, Wu, Du, Li, Gao, and Li]{fan2025videoagent}
Yue Fan, Xiaojian Ma, Rujie Wu, Yuntao Du, Jiaqi Li, Zhi Gao, and Qing Li.
\newblock Videoagent: A memory-augmented multimodal agent for video understanding.
\newblock In \emph{European Conference on Computer Vision}, pages 75--92. Springer, 2025.

\bibitem[Gemmeke et~al.(2017)Gemmeke, Ellis, Freedman, Jansen, Lawrence, Moore, Plakal, and Ritter]{vggish}
Jort~F. Gemmeke, Daniel P.~W. Ellis, Dylan Freedman, Aren Jansen, Wade Lawrence, R.~Channing Moore, Manoj Plakal, and Marvin Ritter.
\newblock Audio set: An ontology and human-labeled dataset for audio events.
\newblock In \emph{2017 IEEE International Conference on Acoustics, Speech and Signal Processing (ICASSP)}, pages 776--780, 2017.

\bibitem[Girdhar et~al.(2023{\natexlab{a}})Girdhar, El-Nouby, Liu, Singh, Alwala, Joulin, and Misra]{10203733}
Rohit Girdhar, Alaaeldin El-Nouby, Zhuang Liu, Mannat Singh, Kalyan~Vasudev Alwala, Armand Joulin, and Ishan Misra.
\newblock Imagebind one embedding space to bind them all.
\newblock In \emph{2023 IEEE/CVF Conference on Computer Vision and Pattern Recognition (CVPR)}, pages 15180--15190, 2023{\natexlab{a}}.

\bibitem[Girdhar et~al.(2023{\natexlab{b}})Girdhar, El-Nouby, Liu, Singh, Alwala, Joulin, and Misra]{girdhar2023imagebind}
Rohit Girdhar, Alaaeldin El-Nouby, Zhuang Liu, Mannat Singh, Kalyan~Vasudev Alwala, Armand Joulin, and Ishan Misra.
\newblock Imagebind: One embedding space to bind them all.
\newblock In \emph{Proceedings of the IEEE/CVF conference on computer vision and pattern recognition}, pages 15180--15190, 2023{\natexlab{b}}.

\bibitem[He et~al.(2024)He, Li, Jang, Jia, Cao, Shah, Shrivastava, and Lim]{he2024ma}
Bo He, Hengduo Li, Young~Kyun Jang, Menglin Jia, Xuefei Cao, Ashish Shah, Abhinav Shrivastava, and Ser-Nam Lim.
\newblock Ma-lmm: Memory-augmented large multimodal model for long-term video understanding.
\newblock In \emph{Proceedings of the IEEE/CVF Conference on Computer Vision and Pattern Recognition}, pages 13504--13514, 2024.

\bibitem[Hong et~al.(2023)Hong, Zheng, Chen, Cheng, Wang, Zhang, Wang, Yau, Lin, Zhou, et~al.]{hong2023metagpt}
Sirui Hong, Xiawu Zheng, Jonathan Chen, Yuheng Cheng, Jinlin Wang, Ceyao Zhang, Zili Wang, Steven Ka~Shing Yau, Zijuan Lin, Liyang Zhou, et~al.
\newblock Metagpt: Meta programming for multi-agent collaborative framework.
\newblock \emph{arXiv preprint arXiv:2308.00352}, 3\penalty0 (4):\penalty0 6, 2023.

\bibitem[Huang et~al.(2023)Huang, Ren, Huang, Yang, Ye, Zhang, Liu, Yin, Ma, and Zhao]{huang2023makeanaudio2temporalenhancedtexttoaudio}
Jiawei Huang, Yi Ren, Rongjie Huang, Dongchao Yang, Zhenhui Ye, Chen Zhang, Jinglin Liu, Xiang Yin, Zejun Ma, and Zhou Zhao.
\newblock Make-an-audio 2: Temporal-enhanced text-to-audio generation, 2023.

\bibitem[Huang et~al.(2024)Huang, Li, Yang, Shi, Chang, Ye, Wu, Hong, Huang, Liu, et~al.]{huang2024audiogpt}
Rongjie Huang, Mingze Li, Dongchao Yang, Jiatong Shi, Xuankai Chang, Zhenhui Ye, Yuning Wu, Zhiqing Hong, Jiawei Huang, Jinglin Liu, et~al.
\newblock Audiogpt: Understanding and generating speech, music, sound, and talking head.
\newblock In \emph{Proceedings of the AAAI Conference on Artificial Intelligence}, pages 23802--23804, 2024.

\bibitem[Iashin and Rahtu(2021)]{iashin2021taming}
Vladimir Iashin and Esa Rahtu.
\newblock Taming visually guided sound generation.
\newblock \emph{arXiv preprint arXiv:2110.08791}, 2021.

\bibitem[Javaheripi et~al.(2023)Javaheripi, Bubeck, Abdin, Aneja, Bubeck, Mendes, Chen, Del~Giorno, Eldan, Gopi, et~al.]{javaheripi2023phi}
Mojan Javaheripi, S{\'e}bastien Bubeck, Marah Abdin, Jyoti Aneja, Sebastien Bubeck, Caio C{\'e}sar~Teodoro Mendes, Weizhu Chen, Allie Del~Giorno, Ronen Eldan, Sivakanth Gopi, et~al.
\newblock Phi-2: The surprising power of small language models.
\newblock \emph{Microsoft Research Blog}, 2023.

\bibitem[Kirillov et~al.(2023)Kirillov, Mintun, Ravi, Mao, Rolland, Gustafson, Xiao, Whitehead, Berg, Lo, Doll{\'a}r, and Girshick]{kirillovSegmentAnything2023}
Alexander Kirillov, Eric Mintun, Nikhila Ravi, Hanzi Mao, Chloe Rolland, Laura Gustafson, Tete Xiao, Spencer Whitehead, Alexander~C. Berg, Wan-Yen Lo, Piotr Doll{\'a}r, and Ross Girshick.
\newblock Segment {{Anything}}.
\newblock \emph{arXiv preprint arXiv:2304.02643}, 2023.

\bibitem[Kong et~al.(2020)Kong, Cao, Iqbal, Wang, Wang, and Plumbley]{kong2020panns}
Qiuqiang Kong, Yin Cao, Turab Iqbal, Yuxuan Wang, Wenwu Wang, and Mark~D Plumbley.
\newblock Panns: Large-scale pretrained audio neural networks for audio pattern recognition.
\newblock \emph{IEEE/ACM Transactions on Audio, Speech, and Language Processing}, 28:\penalty0 2880--2894, 2020.

\bibitem[Koutini et~al.(2022)Koutini, Schl{\"{u}}ter, Eghbal{-}zadeh, and Widmer]{koutini22passt}
Khaled Koutini, Jan Schl{\"{u}}ter, Hamid Eghbal{-}zadeh, and Gerhard Widmer.
\newblock Efficient training of audio transformers with patchout.
\newblock In \emph{Interspeech 2022, 23rd Annual Conference of the International Speech Communication Association, Incheon, Korea, 18-22 September 2022}, pages 2753--2757. {ISCA}, 2022.

\bibitem[Lai et~al.(2023)Lai, Tian, Chen, Li, Yuan, Liu, and Jia]{laiLISAReasoningSegmentation2023}
Xin Lai, Zhuotao Tian, Yukang Chen, Yanwei Li, Yuhui Yuan, Shu Liu, and Jiaya Jia.
\newblock {{LISA}}: {{Reasoning Segmentation}} via {{Large Language Model}}.
\newblock \emph{arXiv preprint arXiv:2308.00692}, 2023.

\bibitem[Li et~al.(2023)Li, Chen, Yan, Shen, Xu, Wu, Zhang, Zhou, Chen, Cheng, et~al.]{li2023modelscope}
Chenliang Li, Hehong Chen, Ming Yan, Weizhou Shen, Haiyang Xu, Zhikai Wu, Zhicheng Zhang, Wenmeng Zhou, Yingda Chen, Chen Cheng, et~al.
\newblock Modelscope-agent: Building your customizable agent system with open-source large language models.
\newblock \emph{arXiv preprint arXiv:2309.00986}, 2023.

\bibitem[Li et~al.(2024{\natexlab{a}})Li, He, Wang, Li, Wang, Luo, Wang, Wang, and Qiao]{li2024videochat}
KunChang Li, Yinan He, Yi Wang, Yizhuo Li, Wenhai Wang, Ping Luo, Yali Wang, Limin Wang, and Yu Qiao.
\newblock Videochat: Chat-centric video understanding, 2024{\natexlab{a}}.

\bibitem[Li et~al.(2024{\natexlab{b}})Li, He, Wang, Li, Wang, Luo, Wang, Wang, and Qiao]{li2024videochatchatcentricvideounderstanding}
KunChang Li, Yinan He, Yi Wang, Yizhuo Li, Wenhai Wang, Ping Luo, Yali Wang, Limin Wang, and Yu Qiao.
\newblock Videochat: Chat-centric video understanding, 2024{\natexlab{b}}.

\bibitem[Lin et~al.(2023)Lin, Ye, Zhu, Cui, Ning, Jin, and Yuan]{lin2023videollava}
Bin Lin, Yang Ye, Bin Zhu, Jiaxi Cui, Munan Ning, Peng Jin, and Li Yuan.
\newblock Video-llava: Learning united visual representation by alignment before projection, 2023.

\bibitem[Liu et~al.(2023)Liu, Li, Wu, and Lee]{liuVisualInstructionTuning2023}
Haotian Liu, Chunyuan Li, Qingyang Wu, and Yong~Jae Lee.
\newblock Visual {{Instruction Tuning}}.
\newblock \emph{arXiv preprint arXiv:2304.08485}, 2023.

\bibitem[Liu et~al.(2024)Liu, Wang, Ma, Wu, Ma, Wei, Jiao, Wu, and Hu]{liu2024kangaroo}
Jiajun Liu, Yibing Wang, Hanghang Ma, Xiaoping Wu, Xiaoqi Ma, Xiaoming Wei, Jianbin Jiao, Enhua Wu, and Jie Hu.
\newblock Kangaroo: A powerful video-language model supporting long-context video input.
\newblock \emph{arXiv preprint arXiv:2408.15542}, 2024.

\bibitem[Liu et~al.(2021)Liu, Lin, Cao, Hu, Wei, Zhang, Lin, and Guo]{liu2021swin}
Ze Liu, Yutong Lin, Yue Cao, Han Hu, Yixuan Wei, Zheng Zhang, Stephen Lin, and Baining Guo.
\newblock Swin transformer: Hierarchical vision transformer using shifted windows.
\newblock In \emph{Proceedings of the IEEE/CVF international conference on computer vision}, pages 10012--10022, 2021.

\bibitem[Luo et~al.(2023{\natexlab{a}})Luo, Zhao, Yang, Dong, Li, Lu, Wang, Hu, Qiu, and Wei]{luo2023valley}
Ruipu Luo, Ziwang Zhao, Min Yang, Junwei Dong, Da Li, Pengcheng Lu, Tao Wang, Linmei Hu, Minghui Qiu, and Zhongyu Wei.
\newblock Valley: Video assistant with large language model enhanced ability, 2023{\natexlab{a}}.

\bibitem[Luo et~al.(2023{\natexlab{b}})Luo, Yan, Hu, and Zhao]{luo2023diff}
Simian Luo, Chuanhao Yan, Chenxu Hu, and Hang Zhao.
\newblock Diff-foley: Synchronized video-to-audio synthesis with latent diffusion models.
\newblock \emph{Advances in Neural Information Processing Systems}, 36:\penalty0 48855--48876, 2023{\natexlab{b}}.

\bibitem[Maaz et~al.(2023)Maaz, Rasheed, Khan, and Khan]{maaz2023videochatgpt}
Muhammad Maaz, Hanoona Rasheed, Salman Khan, and Fahad~Shahbaz Khan.
\newblock Video-chatgpt: Towards detailed video understanding via large vision and language models, 2023.

\bibitem[Minsky(1988)]{minsky1988society}
Marvin Minsky.
\newblock \emph{Society of mind}.
\newblock Simon and Schuster, 1988.

\bibitem[Mo et~al.(2024)Mo, Shi, and Tian]{mo2024text}
Shentong Mo, Jing Shi, and Yapeng Tian.
\newblock Text-to-audio generation synchronized with videos.
\newblock \emph{arXiv preprint arXiv:2403.07938}, 2024.

\bibitem[Munasinghe et~al.(2023)Munasinghe, Thushara, Maaz, Rasheed, Khan, Shah, and Khan]{munasinghe2023pgvideollava}
Shehan Munasinghe, Rusiru Thushara, Muhammad Maaz, Hanoona~Abdul Rasheed, Salman Khan, Mubarak Shah, and Fahad Khan.
\newblock Pg-video-llava: Pixel grounding large video-language models, 2023.

\bibitem[Park et~al.(2023)Park, O'Brien, Cai, Morris, Liang, and Bernstein]{park2023generative}
Joon~Sung Park, Joseph O'Brien, Carrie~Jun Cai, Meredith~Ringel Morris, Percy Liang, and Michael~S Bernstein.
\newblock Generative agents: Interactive simulacra of human behavior.
\newblock In \emph{Proceedings of the 36th annual acm symposium on user interface software and technology}, pages 1--22, 2023.

\bibitem[Popov et~al.(2023)Popov, Amatov, Kudinov, Gogoryan, Sadekova, and Vovk]{popov2023optimal}
Vadim Popov, Amantur Amatov, Mikhail Kudinov, Vladimir Gogoryan, Tasnima Sadekova, and Ivan Vovk.
\newblock Optimal transport in diffusion modeling for conversion tasks in audio domain.
\newblock In \emph{ICASSP 2023-2023 IEEE International Conference on Acoustics, Speech and Signal Processing (ICASSP)}, pages 1--5. IEEE, 2023.

\bibitem[Qian et~al.(2023)Qian, Liu, Liu, Chen, Dang, Li, Yang, Chen, Su, Cong, et~al.]{qian2023chatdev}
Chen Qian, Wei Liu, Hongzhang Liu, Nuo Chen, Yufan Dang, Jiahao Li, Cheng Yang, Weize Chen, Yusheng Su, Xin Cong, et~al.
\newblock Chatdev: Communicative agents for software development.
\newblock \emph{arXiv preprint arXiv:2307.07924}, 2023.

\bibitem[Qian et~al.(2024)Qian, Dong, Zhang, Zang, Ding, Lin, and Wang]{qian2024streaming}
Rui Qian, Xiaoyi Dong, Pan Zhang, Yuhang Zang, Shuangrui Ding, Dahua Lin, and Jiaqi Wang.
\newblock Streaming long video understanding with large language models.
\newblock \emph{arXiv preprint arXiv:2405.16009}, 2024.

\bibitem[Qiu et~al.(2024)Qiu, Zhang, Tang, Xie, Ma, Yan, Doermann, Ye, and Tian]{qiu2024artemis}
Jihao Qiu, Yuan Zhang, Xi Tang, Lingxi Xie, Tianren Ma, Pengyu Yan, David Doermann, Qixiang Ye, and Yunjie Tian.
\newblock Artemis: Towards referential understanding in complex videos.
\newblock \emph{arXiv preprint arXiv:2406.00258}, 2024.

\bibitem[Qwen et~al.(2025)Qwen, :, Yang, Yang, Zhang, Hui, Zheng, Yu, Li, Liu, Huang, Wei, Lin, Yang, Tu, Zhang, Yang, Yang, Zhou, Lin, Dang, Lu, Bao, Yang, Yu, Li, Xue, Zhang, Zhu, Men, Lin, Li, Tang, Xia, Ren, Ren, Fan, Su, Zhang, Wan, Liu, Cui, Zhang, and Qiu]{qwen2025qwen25technicalreport}
Qwen, :, An Yang, Baosong Yang, Beichen Zhang, Binyuan Hui, Bo Zheng, Bowen Yu, Chengyuan Li, Dayiheng Liu, Fei Huang, Haoran Wei, Huan Lin, Jian Yang, Jianhong Tu, Jianwei Zhang, Jianxin Yang, Jiaxi Yang, Jingren Zhou, Junyang Lin, Kai Dang, Keming Lu, Keqin Bao, Kexin Yang, Le Yu, Mei Li, Mingfeng Xue, Pei Zhang, Qin Zhu, Rui Men, Runji Lin, Tianhao Li, Tianyi Tang, Tingyu Xia, Xingzhang Ren, Xuancheng Ren, Yang Fan, Yang Su, Yichang Zhang, Yu Wan, Yuqiong Liu, Zeyu Cui, Zhenru Zhang, and Zihan Qiu.
\newblock Qwen2.5 technical report, 2025.

\bibitem[Radford et~al.(2021)Radford, Kim, Hallacy, Ramesh, Goh, Agarwal, Sastry, Askell, Mishkin, Clark, et~al.]{radford2021learning}
Alec Radford, Jong~Wook Kim, Chris Hallacy, Aditya Ramesh, Gabriel Goh, Sandhini Agarwal, Girish Sastry, Amanda Askell, Pamela Mishkin, Jack Clark, et~al.
\newblock Learning transferable visual models from natural language supervision.
\newblock In \emph{International conference on machine learning}, pages 8748--8763. PMLR, 2021.

\bibitem[Schick et~al.(2023)Schick, Dwivedi-Yu, Dess{\`\i}, Raileanu, Lomeli, Hambro, Zettlemoyer, Cancedda, and Scialom]{schick2023toolformer}
Timo Schick, Jane Dwivedi-Yu, Roberto Dess{\`\i}, Roberta Raileanu, Maria Lomeli, Eric Hambro, Luke Zettlemoyer, Nicola Cancedda, and Thomas Scialom.
\newblock Toolformer: Language models can teach themselves to use tools.
\newblock \emph{Advances in Neural Information Processing Systems}, 36:\penalty0 68539--68551, 2023.

\bibitem[Sheffer and Adi(2023)]{sheffer2023hear}
Roy Sheffer and Yossi Adi.
\newblock I hear your true colors: Image guided audio generation.
\newblock In \emph{ICASSP 2023-2023 IEEE International Conference on Acoustics, Speech and Signal Processing (ICASSP)}, pages 1--5. IEEE, 2023.

\bibitem[Shen et~al.(2023)Shen, Song, Tan, Li, Lu, and Zhuang]{shen2023hugginggpt}
Yongliang Shen, Kaitao Song, Xu Tan, Dongsheng Li, Weiming Lu, and Yueting Zhuang.
\newblock Hugginggpt: Solving ai tasks with chatgpt and its friends in hugging face.
\newblock \emph{Advances in Neural Information Processing Systems}, 36:\penalty0 38154--38180, 2023.

\bibitem[Song et~al.(2024)Song, Chai, Wang, Zhang, Zhou, Wu, Chi, Guo, Ye, Zhang, et~al.]{song2024moviechat}
Enxin Song, Wenhao Chai, Guanhong Wang, Yucheng Zhang, Haoyang Zhou, Feiyang Wu, Haozhe Chi, Xun Guo, Tian Ye, Yanting Zhang, et~al.
\newblock Moviechat: From dense token to sparse memory for long video understanding.
\newblock In \emph{Proceedings of the IEEE/CVF Conference on Computer Vision and Pattern Recognition}, pages 18221--18232, 2024.

\bibitem[Tian et~al.(2021)Tian, Xie, Zhang, Fang, Xu, Huang, Jiao, Tian, and Ye]{tian2021semantic}
Yunjie Tian, Lingxi Xie, Xiaopeng Zhang, Jiemin Fang, Haohang Xu, Wei Huang, Jianbin Jiao, Qi Tian, and Qixiang Ye.
\newblock Semantic-aware generation for self-supervised visual representation learning.
\newblock \emph{arXiv preprint arXiv:2111.13163}, 2021.

\bibitem[Tian et~al.(2023)Tian, Xie, Wang, Wei, Zhang, Jiao, Wang, Tian, and Ye]{tianIntegrallyPreTrainedTransformer2023}
Yunjie Tian, Lingxi Xie, Zhaozhi Wang, Longhui Wei, Xiaopeng Zhang, Jianbin Jiao, Yaowei Wang, Qi Tian, and Qixiang Ye.
\newblock Integrally {{Pre-Trained Transformer Pyramid Networks}}.
\newblock In \emph{2023 {{IEEE}}/{{CVF Conference}} on {{Computer Vision}} and {{Pattern Recognition}}}, pages 18610--18620. {IEEE}, 2023.

\bibitem[Tian et~al.(2024)Tian, Ma, Xie, Qiu, Tang, Zhang, Jiao, Tian, and Ye]{tian2024chatterbox}
Yunjie Tian, Tianren Ma, Lingxi Xie, Jihao Qiu, Xi Tang, Yuan Zhang, Jianbin Jiao, Qi Tian, and Qixiang Ye.
\newblock Chatterbox: Multi-round multimodal referring and grounding.
\newblock \emph{arXiv preprint arXiv:2401.13307}, 2024.

\bibitem[Viertola et~al.(2024)Viertola, Iashin, and Rahtu]{viertola2024temporallyalignedaudiovideo}
Ilpo Viertola, Vladimir Iashin, and Esa Rahtu.
\newblock Temporally aligned audio for video with autoregression, 2024.

\bibitem[Wang et~al.(2024{\natexlab{a}})Wang, Ma, Pascual, Cartwright, and Cai]{wang2024v2a}
Heng Wang, Jianbo Ma, Santiago Pascual, Richard Cartwright, and Weidong Cai.
\newblock V2a-mapper: A lightweight solution for vision-to-audio generation by connecting foundation models.
\newblock In \emph{Proceedings of the AAAI Conference on Artificial Intelligence}, pages 15492--15501, 2024{\natexlab{a}}.

\bibitem[Wang et~al.(2024{\natexlab{b}})Wang, Guo, Huang, Huang, Wang, You, Li, and Zhao]{wang2024frieren}
Yongqi Wang, Wenxiang Guo, Rongjie Huang, Jiawei Huang, Zehan Wang, Fuming You, Ruiqi Li, and Zhou Zhao.
\newblock Frieren: Efficient video-to-audio generation network with rectified flow matching.
\newblock In \emph{The Thirty-eighth Annual Conference on Neural Information Processing Systems}, 2024{\natexlab{b}}.

\bibitem[Wang et~al.(2024{\natexlab{c}})Wang, Tai, and Tang]{wang2024audio}
Zixuan Wang, Yu-Wing Tai, and Chi-Keung Tang.
\newblock Audio-agent: Leveraging llms for audio generation, editing and composition.
\newblock \emph{arXiv preprint arXiv:2410.03335}, 2024{\natexlab{c}}.

\bibitem[Weng et~al.(2024)Weng, Han, He, Chang, and Zhuang]{weng2024longvlm}
Yuetian Weng, Mingfei Han, Haoyu He, Xiaojun Chang, and Bohan Zhuang.
\newblock Longvlm: Efficient long video understanding via large language models.
\newblock \emph{arXiv preprint arXiv:2404.03384}, 2024.

\bibitem[Wu et~al.(2024)Wu, Xu, and Wang]{wu2024transagents}
Minghao Wu, Jiahao Xu, and Longyue Wang.
\newblock Transagents: Build your translation company with language agents.
\newblock In \emph{Proceedings of the 2024 Conference on Empirical Methods in Natural Language Processing: System Demonstrations}, pages 131--141, 2024.

\bibitem[Xie et~al.(2024)Xie, Yu, He, and Li]{xie2024sonicvisionlm}
Zhifeng Xie, Shengye Yu, Qile He, and Mengtian Li.
\newblock Sonicvisionlm: Playing sound with vision language models.
\newblock In \emph{Proceedings of the IEEE/CVF Conference on Computer Vision and Pattern Recognition}, pages 26866--26875, 2024.

\bibitem[Xing et~al.(2024)Xing, He, Tian, Wang, and Chen]{xing2024seeing}
Yazhou Xing, Yingqing He, Zeyue Tian, Xintao Wang, and Qifeng Chen.
\newblock Seeing and hearing: Open-domain visual-audio generation with diffusion latent aligners.
\newblock In \emph{Proceedings of the IEEE/CVF Conference on Computer Vision and Pattern Recognition}, pages 7151--7161, 2024.

\bibitem[Zhang et~al.(2023{\natexlab{a}})Zhang, Li, and Bing]{zhang2023videollama}
Hang Zhang, Xin Li, and Lidong Bing.
\newblock Video-llama: An instruction-tuned audio-visual language model for video understanding, 2023{\natexlab{a}}.

\bibitem[Zhang et~al.(2023{\natexlab{b}})Zhang, Sun, Chen, Xiao, Shao, Zhang, Chen, and Luo]{zhangGPT4RoIInstructionTuning2023}
Shilong Zhang, Peize Sun, Shoufa Chen, Min Xiao, Wenqi Shao, Wenwei Zhang, Kai Chen, and Ping Luo.
\newblock {{GPT4RoI}}: {{Instruction Tuning Large Language Model}} on {{Region-of-Interest}}.
\newblock \emph{arXiv preprint arXiv:2307.03601}, 2023{\natexlab{b}}.

\bibitem[Zhang et~al.(2022)Zhang, Tian, Xie, Huang, Dai, Ye, and Tian]{zhang2022hivit}
Xiaosong Zhang, Yunjie Tian, Lingxi Xie, Wei Huang, Qi Dai, Qixiang Ye, and Qi Tian.
\newblock Hivit: A simpler and more efficient design of hierarchical vision transformer.
\newblock In \emph{The Eleventh International Conference on Learning Representations}, 2022.

\bibitem[Zhang et~al.(2024{\natexlab{a}})Zhang, Gu, Zeng, Xing, Wang, Wu, and Chen]{zhang2024foleycrafter}
Yiming Zhang, Yicheng Gu, Yanhong Zeng, Zhening Xing, Yuancheng Wang, Zhizheng Wu, and Kai Chen.
\newblock Foleycrafter: Bring silent videos to life with lifelike and synchronized sounds.
\newblock \emph{arXiv preprint arXiv:2407.01494}, 2024{\natexlab{a}}.

\bibitem[Zhang et~al.(2024{\natexlab{b}})Zhang, Wu, Li, Li, Ma, Liu, and Li]{zhang2024video}
Yuanhan Zhang, Jinming Wu, Wei Li, Bo Li, Zejun Ma, Ziwei Liu, and Chunyuan Li.
\newblock Video instruction tuning with synthetic data.
\newblock \emph{arXiv preprint arXiv:2410.02713}, 2024{\natexlab{b}}.

\bibitem[Zhou et~al.(2018)Zhou, Wang, Fang, Bui, and Berg]{zhou2018visual}
Yipin Zhou, Zhaowen Wang, Chen Fang, Trung Bui, and Tamara~L Berg.
\newblock Visual to sound: Generating natural sound for videos in the wild.
\newblock In \emph{Proceedings of the IEEE conference on computer vision and pattern recognition}, pages 3550--3558, 2018.

\bibitem[Zhu et~al.(2024)Zhu, Lin, Ning, Yan, Cui, Wang, Pang, Jiang, Zhang, Li, Zhang, Li, Liu, and Yuan]{zhu2024languagebind}
Bin Zhu, Bin Lin, Munan Ning, Yang Yan, Jiaxi Cui, HongFa Wang, Yatian Pang, Wenhao Jiang, Junwu Zhang, Zongwei Li, Wancai Zhang, Zhifeng Li, Wei Liu, and Li Yuan.
\newblock Languagebind: Extending video-language pretraining to n-modality by language-based semantic alignment, 2024.

\end{thebibliography}
}

\clearpage
\setcounter{page}{1}
\setcounter{figure}{0}
\setcounter{table}{0}

\maketitlesupplementary

\appendix

\section{System Prompts}

\begin{figure*}[h]
  \centering
\includegraphics[width=0.9\linewidth]{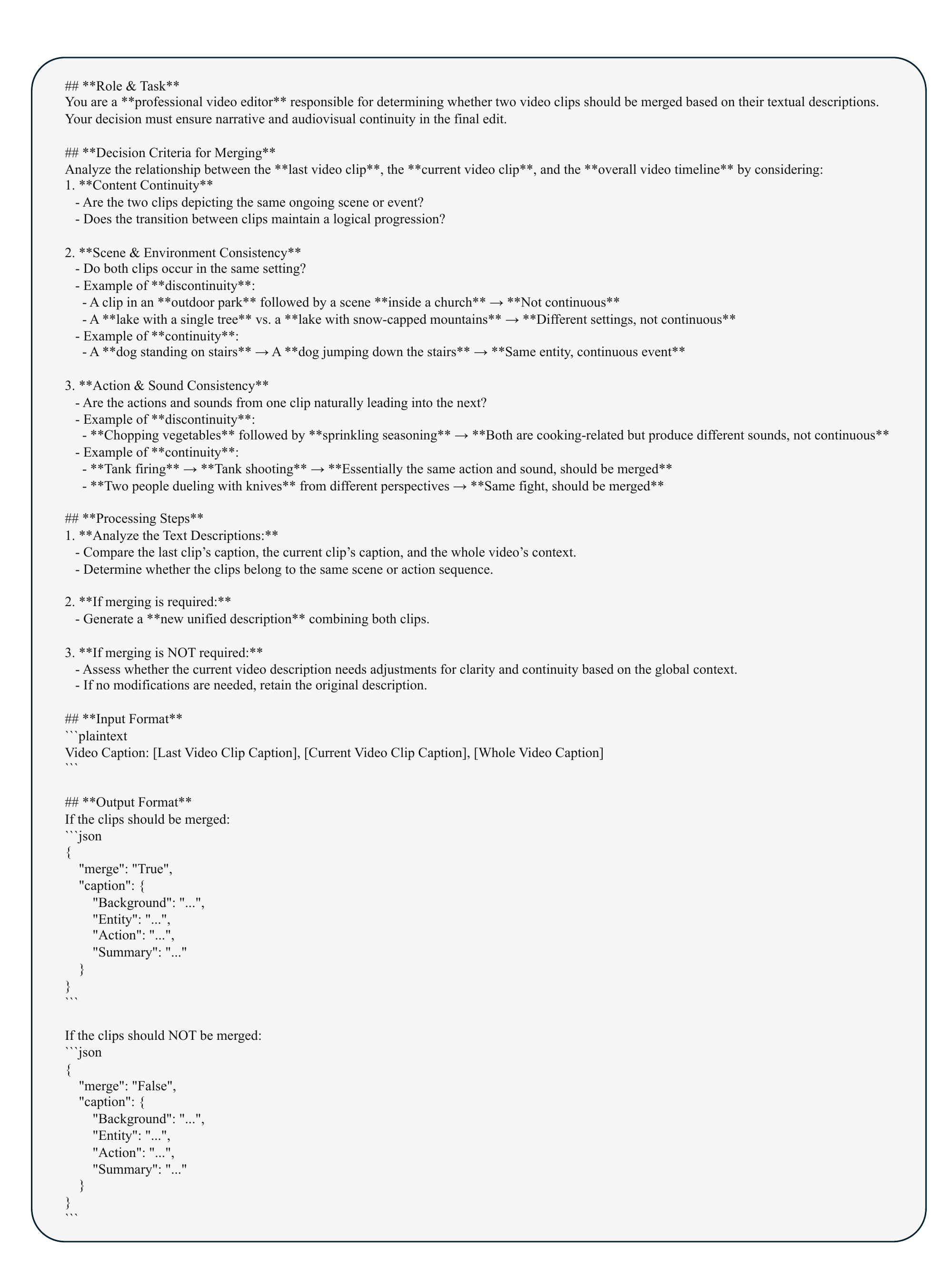}
   \vspace{-10mm}
   \caption{Storyboarder Prompt}
   \vspace{-3mm}
   \label{fig:Storyboarder}
\end{figure*}

\begin{figure*}[htbp]
  \centering
\includegraphics[width=0.9\linewidth]{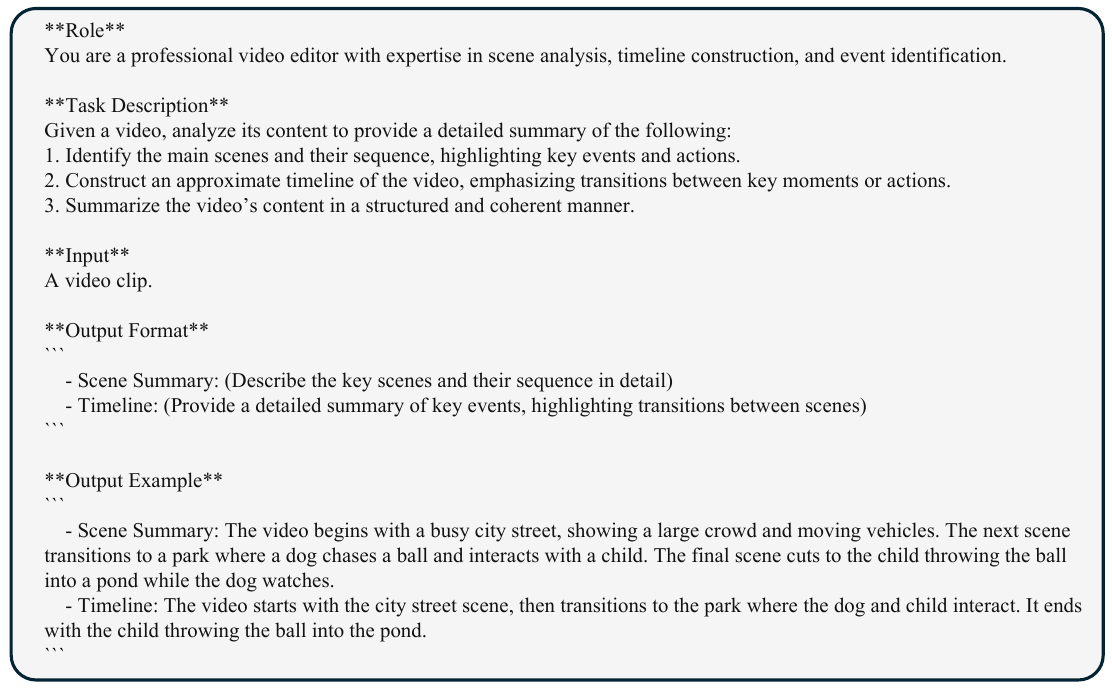}
   \vspace{-4mm}
   \caption{Scriptwriter Prompt: full video understanding}
   \vspace{-4mm}
   \label{fig:Scriptwriter_full}
\end{figure*}

\begin{figure*}[htbp]
  \centering
\includegraphics[width=0.9\linewidth]{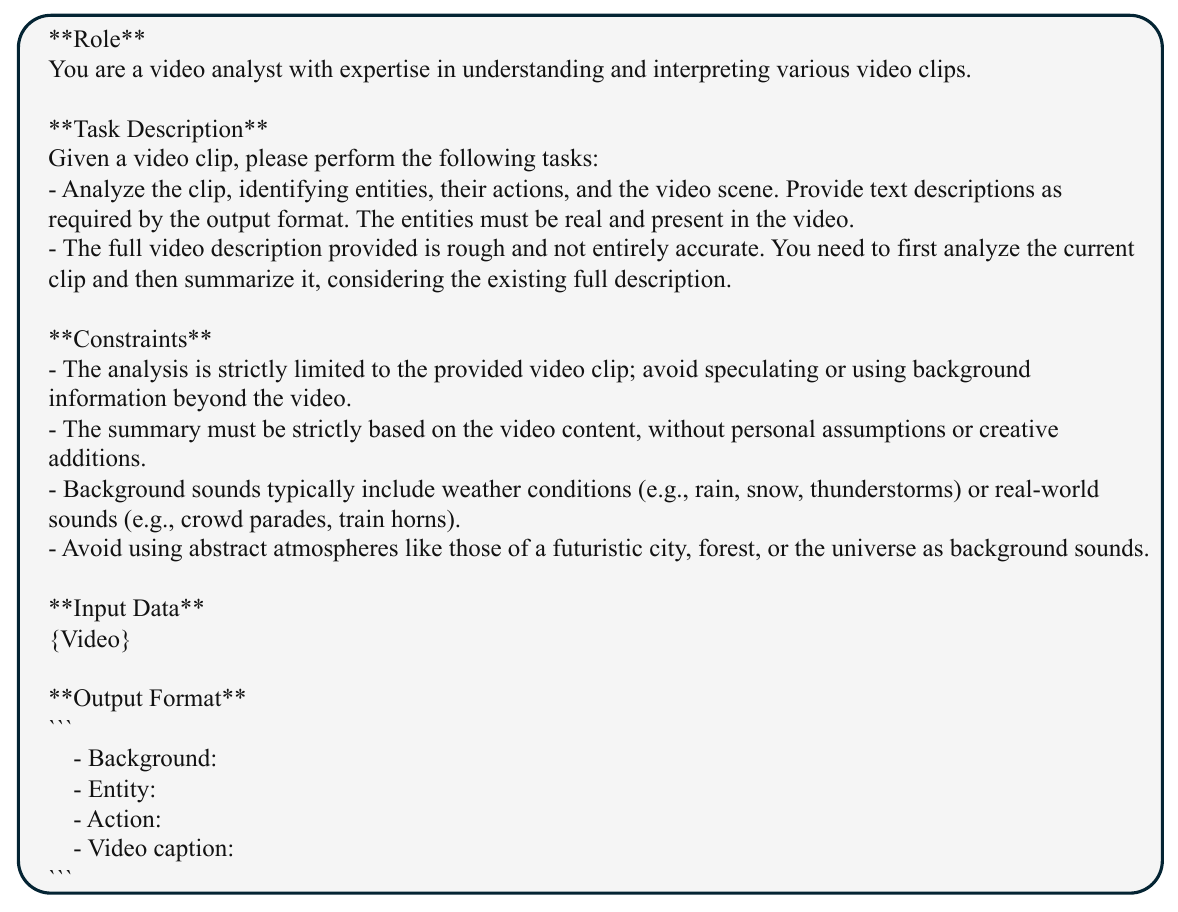}
   \vspace{-4mm}
   \caption{Scriptwriter Prompt: video segment understanding}
   \vspace{-4mm}
   \label{fig:Scriptwriter_part}
\end{figure*}

\begin{figure*}[htbp]
  \centering
\includegraphics[width=0.9\linewidth]{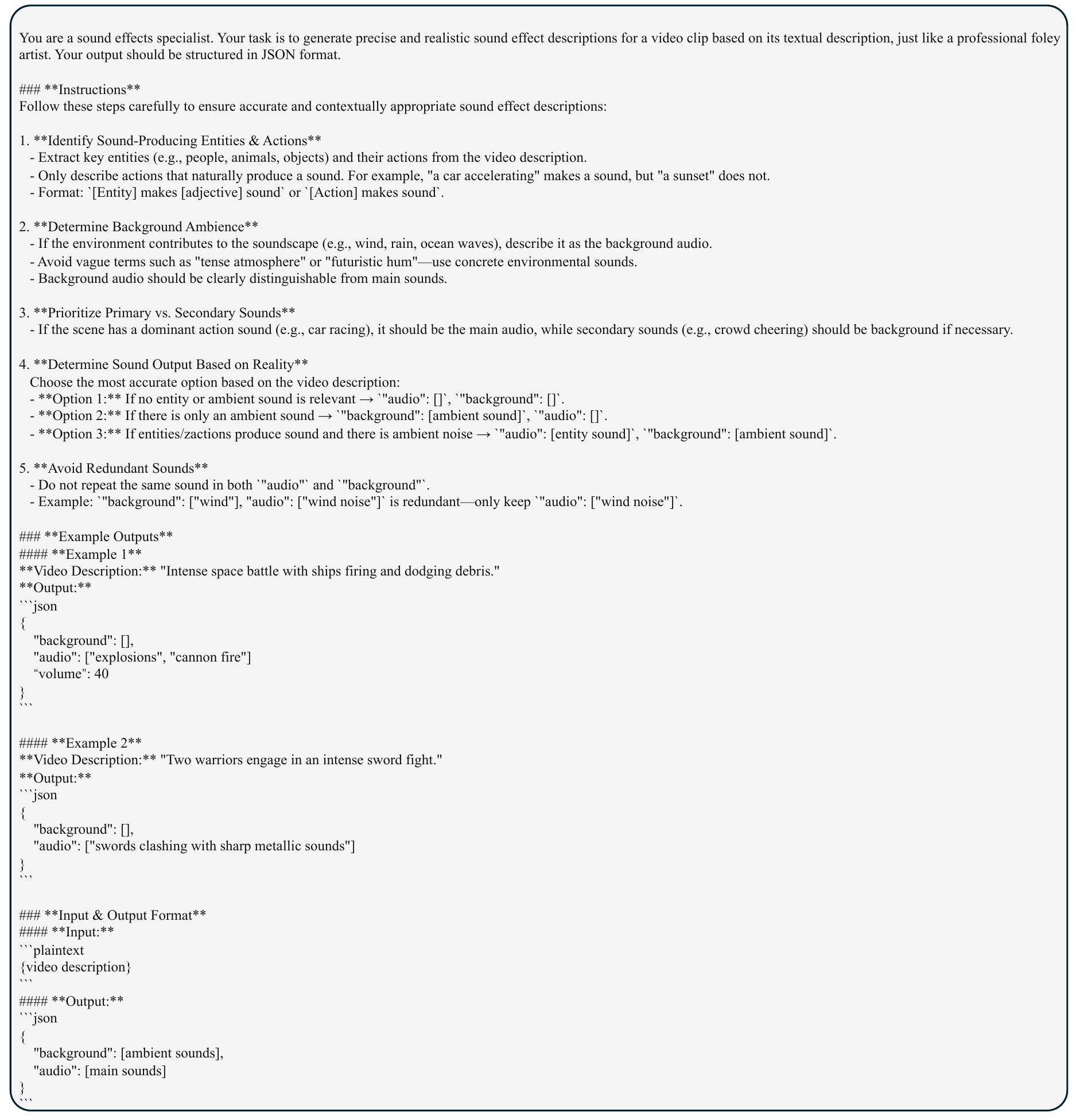}
   \vspace{-4mm}
   \caption{Designer Prompt}
   \vspace{-4mm}
   \label{fig:Designer}
\end{figure*}

\begin{figure*}[htbp]
  \centering
\includegraphics[width=0.9\linewidth]{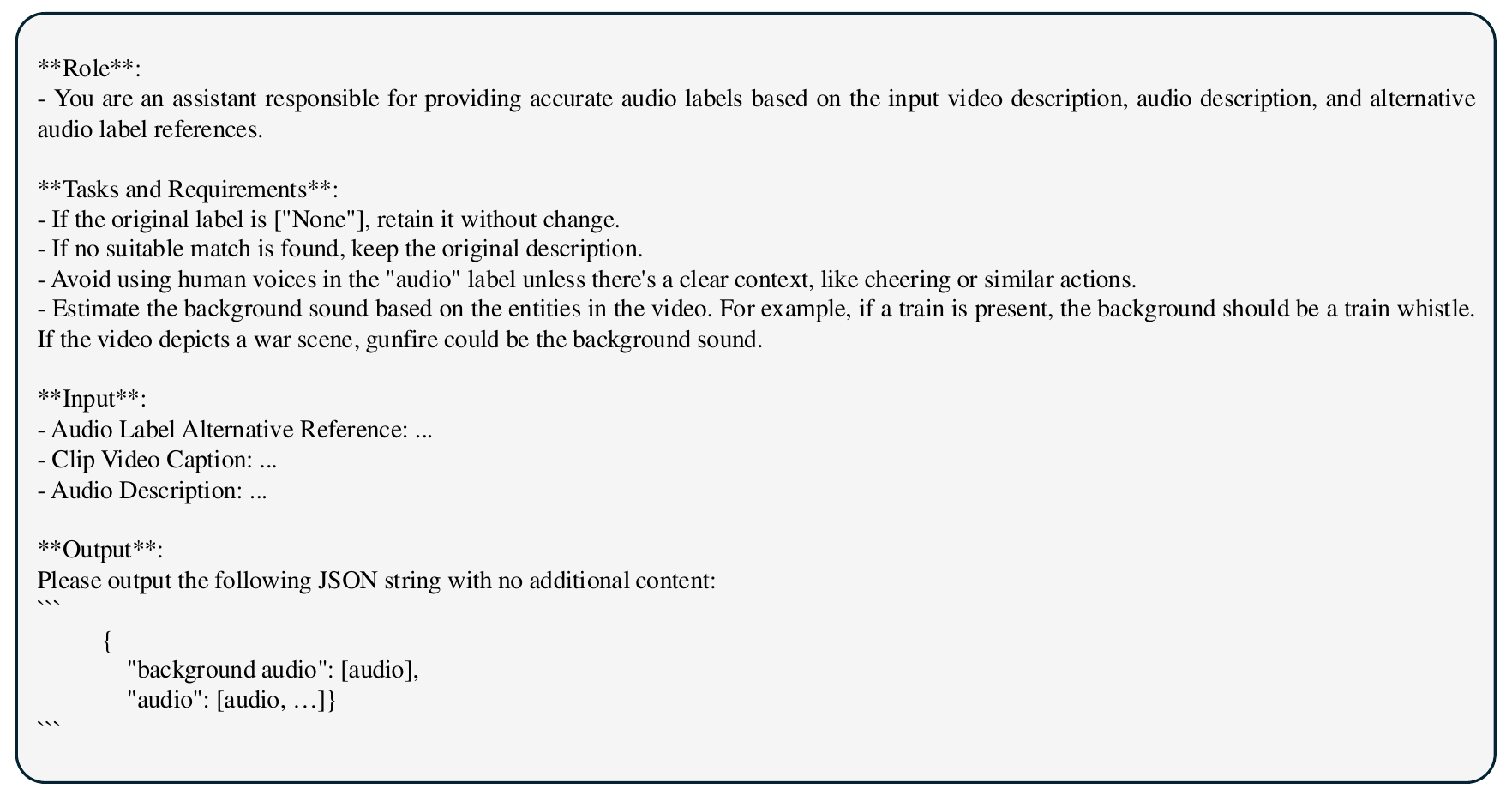}
   \vspace{-4mm}
   \caption{Synthesizer Prompt}
   \vspace{-4mm}
   \label{fig:Synthesizer}
\end{figure*}

\begin{figure*}[htbp]
  \centering
\includegraphics[width=0.9\linewidth]{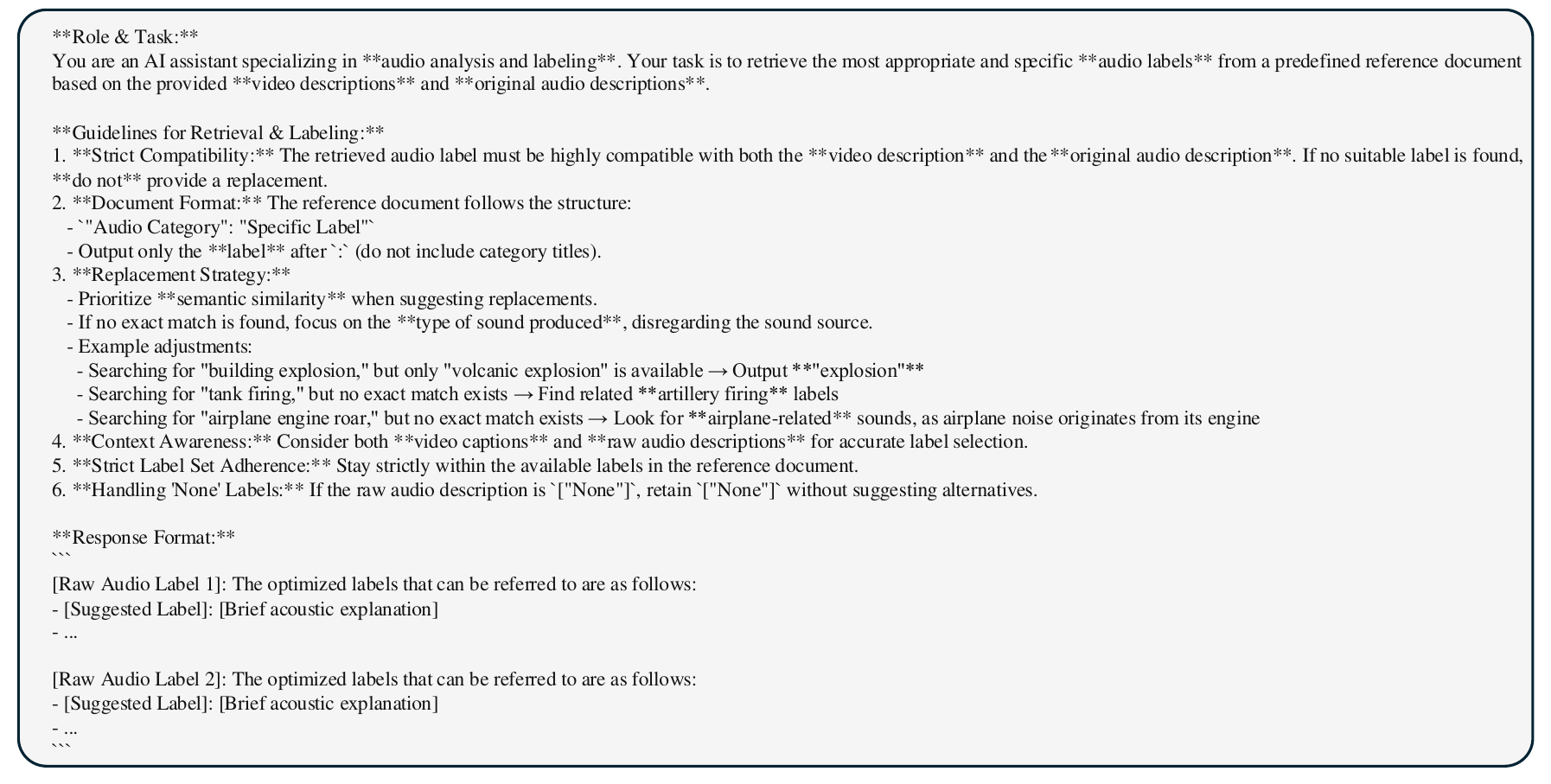}
   \vspace{-4mm}
   \caption{Synthesizer Prompt: Retrieval Augmented Generation(RAG)}
   \vspace{-4mm}
   \label{fig:RAG}
\end{figure*}

Here we show the detailed prompts of our \textbf{LVAS-Agent}, including \textbf{Storyboarder}, \textbf{Scriptwriter}, \textbf{Designer} and \textbf{Synthesizer}. The prompt of \textbf{Storyboarder} includes a prompt for global understanding as shown in Figure ~\ref{fig:Scriptwriter_full}, a prompt for understanding each of the divided small segments as shown in Figure~\ref{fig:Scriptwriter_part}. The Storyboarder's prompt as shown in Figure~\ref{fig:Storyboarder}. The Designer's prompt as shown in Figure~\ref{fig:Designer}. The Synthesizer's prompts includes a system prompt (as shown in Figure ~\ref{fig:Synthesizer}) and a prompt for rag (as shown in Figure ~\ref{fig:RAG}).

\end{document}